%% file: 0-main.tex
\documentclass[conference]{IEEEtran}
\IEEEoverridecommandlockouts

\usepackage[numbers]{natbib}

\usepackage{dsfont} 

\usepackage{flushend}
\usepackage{balance}

\usepackage{hyperref}
\usepackage{graphics}
\usepackage{graphicx}
\usepackage{subcaption}

\usepackage{amsmath}
\usepackage{amssymb}
\usepackage{algorithm}

\usepackage{xcolor}
\usepackage{booktabs}
\usepackage{multirow}
\usepackage{colortbl}

\newcommand{\figref}[1]{Figure\,\ref{fig:#1}}

\newcommand{\secref}[1]{Section\,\ref{sec:#1}}

\newcommand{\tabref}[1]{Table~\ref{tab:#1}}

\hypersetup{draft}

\begin{document}

\title{Learning Task-Oriented Grasping for Tool Manipulation from Simulated Self-Supervision}

\author{
Kuan Fang
\quad Yuke Zhu
\quad Animesh Garg
\quad Andrey Kurenkov
\quad Viraj Mehta
\quad Li Fei-Fei
\quad Silvio Savarese\\
Stanford University, Stanford, CA 94305 USA
}

\maketitle

\begin{abstract}

Tool manipulation is vital for facilitating robots to complete challenging task goals. It requires reasoning about the desired effect of the task and thus properly grasping and manipulating the tool to achieve the task. Task-agnostic grasping optimizes for grasp robustness while ignoring crucial task-specific constraints. In this paper, we propose the Task-Oriented Grasping Network (TOG-Net) to jointly optimize both task-oriented grasping of a tool and the manipulation policy for that tool. The training process of the model is based on large-scale simulated self-supervision with procedurally generated tool objects. We perform both simulated and real-world experiments on two tool-based manipulation tasks: sweeping and hammering. Our model achieves overall 71.1\% task success rate for sweeping and 80.0\% task success rate for hammering. Supplementary material is available at: 
\href{http://bit.ly/task-oriented-grasp}{bit.ly/task-oriented-grasp}.
\end{abstract}

\IEEEpeerreviewmaketitle

\input{1-introduction.tex}

\input{2-related-work.tex}
\input{3-problem-statement.tex}
\input{4-model.tex}
\input{5-dataset.tex}
\input{6-experiment.tex}
\input{7-discussion.tex}

\input{8-acknowledgement.tex}




\input{bibtex/output.bbl}

\end{document}

%% file: 1-introduction.tex
\section{Introduction}
\label{sec:intro}

Tool manipulation can be defined as the employment of a manipulable object, i.e. a tool, to fulfill a task goal. For this purpose, the agent needs to effectively orient and then manipulate the tool so as to achieve the desired effect.
According to Brown et al.~\cite{brown2012tool}, there are four key aspects to learning task-oriented tool usage:
(a) understanding the desired effect, 
(b) identifying properties of an object that make it a suitable tool,
(c) determining the correct orientation of the tool prior to usage, and 
(d) manipulating the tool.
A task-oriented grasp is therefore a grasp that makes it possible to correctly orient the tool and then manipulate it to complete the task. 

\begin{figure}[!t]
\centering
\vspace{-5pt}
  \includegraphics[width=\linewidth]{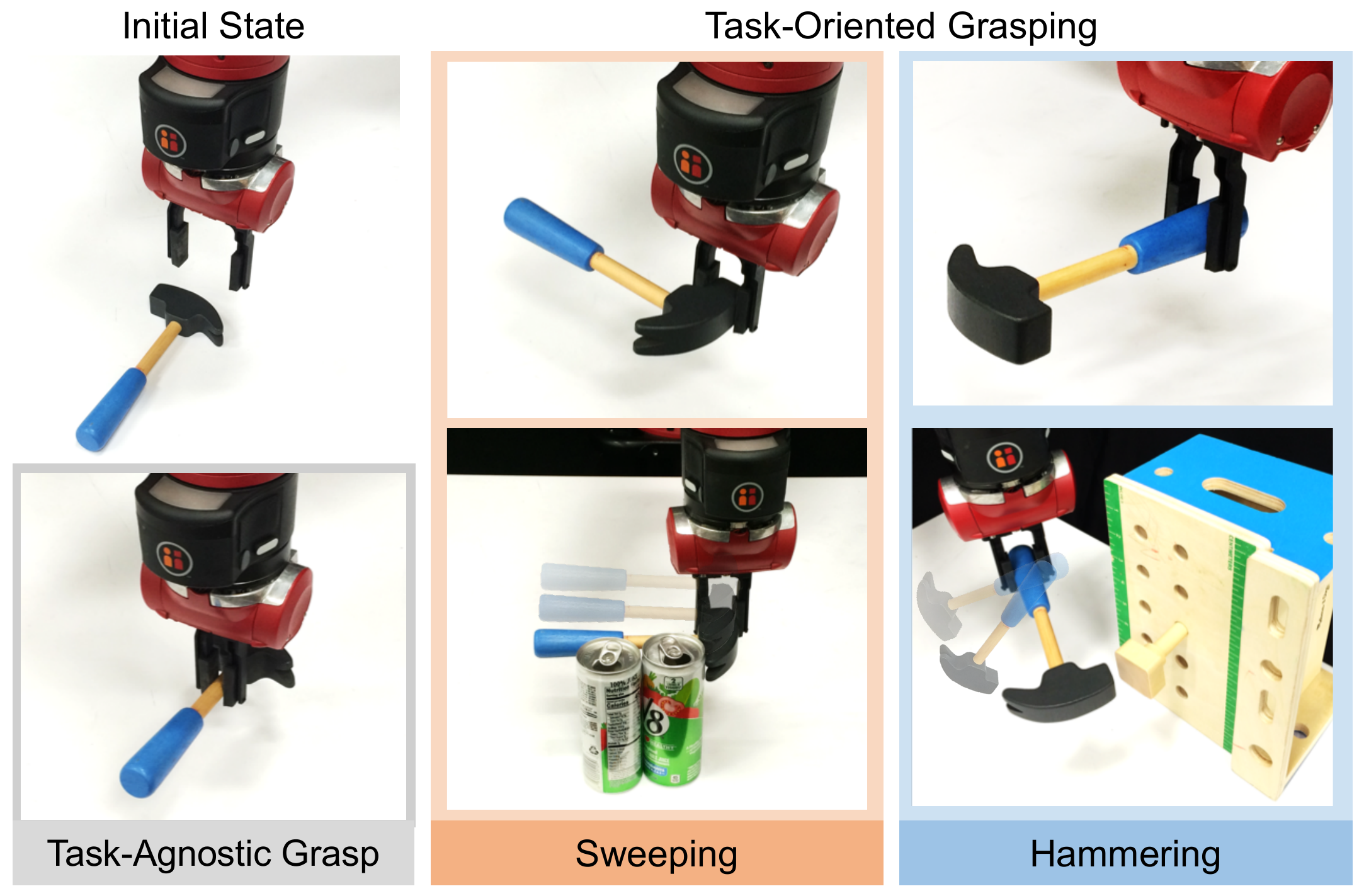}
    \caption{The same object can be grasped by the robot in different ways from the initial state on the table. A task-agnostic grasp can lift up the hammer but it might not be suitable for specific manipulation tasks such as sweeping or hammering. We aim to directly optimize for task success in each episode, by jointly choosing a task-oriented grasp and the subsequent manipulation actions. }

  \label{fig:intro-fig}
  \vspace{-15pt}
\end{figure}

Consider a hammer object as shown in \figref{intro-fig}. The best grasp predicted by a \textit{task-agnostic} task-agnostic grasp prediction model, such as Dex-Net~\cite{mahler2017dex}, is likely to reside close to the center of mass to optimize for robustness. However, the hammering task can be best achieved by holding the hammer at the far end of the handle, thus to generate a high moment at the point of impact on the head. And yet when the same object is used for the sweeping task, it should be grasped by the head since that spares the largest contact surface area with the target objects. In order to optimize for the task success, both grasping robustness and suitability for the manipulation should be considered.


The problem of understanding and using tools has been studied in robotics, computer vision, and psychology~\cite{baber2003cognition,gibson1979affordance,guerin2013survey,huaman2016grasping,osiurak2010grasping,zhu2015understanding}. 
Various studies in robotics have primarily focused on reasoning about the geometric properties of tools~\cite{dang2012semantic,song2010learning}. They often assume prior knowledge of object geometry and require predefined affordance labels and semantic constraints, which has constrained their usefulness in realistic environments with sensory and control uncertainty.
Some pioneering works have also grounded the tool grasping problem in an interactive environment~\cite{mar2015self,mar2017self}. Nonetheless, their approaches relied on hand-engineered feature representations and simplified action spaces, which are limited to the tasks they are tuned to work for and allow only limited generalization to novel objects in complex manipulation tasks. 

In this work, we focus on tool manipulation tasks that consist of two stages. First, the robot picks up a tool resting on the tabletop. Second, it manipulates this tool to complete a task. We propose the Task-Oriented Grasping Network (TOG-Net), a learning-based model for jointly predicting task-oriented grasps and subsequent manipulation actions given the visual inputs. To accommodate the need for large amounts of training data for deep learning, we embrace the self-supervised learning paradigm~\cite{levine2016learning,mar2017self,pinto2016supersizing}, where the robot performs grasping and manipulation attempts and the training labels automatically generated thereby. To scale up our self-supervision data collection, we leverage a real-time physics simulator~\cite{coumans2017bullet} that allows a simulated robot to perform task executions with diverse procedurally generated 3D objects. We evaluate our method on a hammering task and a sweeping task. Our model is proved to learn robust policies that generalize well to novel objects both in simulation and the real world. In the real-world experiments, our model achieves 71.1\% task success rate for sweeping and 80.0\% task success rate for hammering using 9 unseen objects as tools.

Our primary contributions are three-fold:
1) We propose a learning-based model for jointly learning task-oriented grasping and tool manipulation that directly optimizes task success;
2) To train this model, we develop a mechanism for generating large-scale simulated self-supervision with a large repository of procedurally generated 3D objects;
3) We demonstrate that our model can generalize to using novel objects as tools in both simulation and the real world.

%% file: 2-related-work.tex
\section{Related Work}
\label{sec:rw}

\begin{figure*}[!t]
\centering
\vspace{-5pt}
  \includegraphics[width=\linewidth]{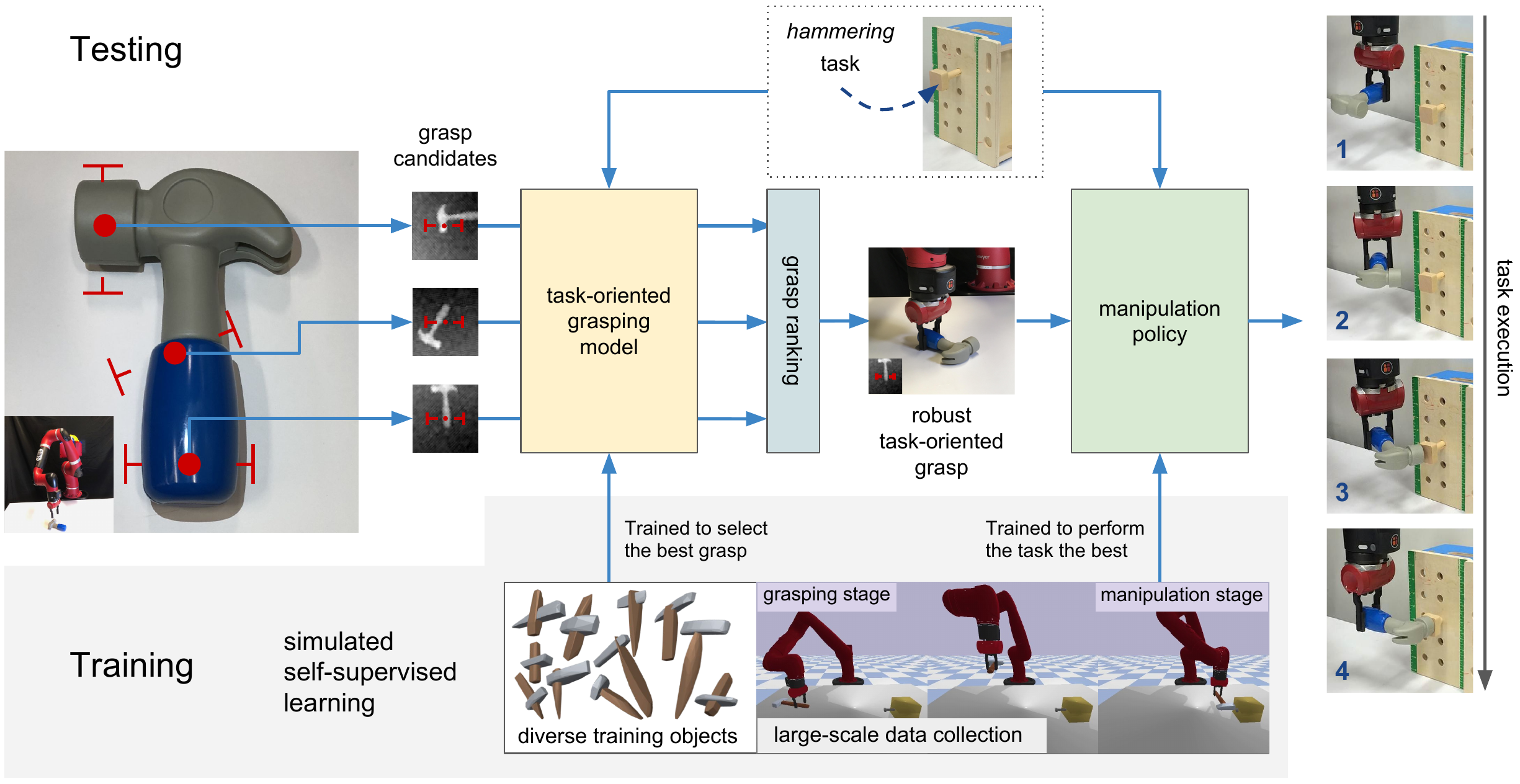}
  \caption{\textbf{Model Overview.} Our model consists of a task-oriented grasping model and a manipulation policy. Given the visual inputs of the object, we sample multiple grasp candidates. The task-oriented grasping model computes a grasp quality score for each candidate based on the planned task, and chooses the grasp with the highest score. Given the observation of the scene, the manipulation policy outputs actions conditioned on the selected grasp. The two modules are trained jointly using simulated self-supervision.}
  \label{fig:overview}
  \vspace{-15pt}
\end{figure*}

\vspace{1mm}
\noindent \textbf{Task-Agnostic Grasping:} Robotic grasping is a long-standing challenge that involves perception and control. Classical methods approach the grasping problem from a purely geometric perspective. They optimize grasp quality measures based on analytic models of geometric constraints, such as force closure and form closure~\cite{ferrari1992planning,rodriguez2012caging,weisz2012pose}. %
While grasp quality measures offer a principled mathematical framework of grasp analysis, their practical usage is limited by the large space of possible grasps. 
Several techniques have been proposed to restrict the search space of grasp candidates. This includes representing objects with shape primitives~\cite{miller2003automatic}, simplifying the search of grasps in a subspace of reduced dimensionality~\cite{ciocarlie2009hand}, and leveraging a dataset of objects with known grasps to speed up grasp selection for novel objects~\cite{goldfeder2009columbia,mahler2016dex}. 
Another limitation of these approaches is that they require the full 3D geometry of the object, which restricts their usage in unstructured real-world environments. The recent development of machine learning techniques, especially deep learning, has enabled a surge of research that applies data-driven methods to robotic grasping~\cite{bohg2014data,kappler2015leveraging}. These learning methods largely fall into two families depending on the source of supervision: 1) supervised learning approaches~\cite{lenz2015deep,mahler2017dex,saxena2008robotic}, where the models are trained with a dataset of objects with ground-truth grasp annotations, and 2) self-supervision approaches~\cite{jang2017end,levine2016learning,pinto2016supersizing,bousmalis2017using,fang2018instance}, where grasp labels are automatically generated by a robot's trial and error on large numbers of real-world or simulated grasp attempts. To address the data-hungry nature of deep neural networks, several works~\cite{mahler2017dex,viereck2017learning} relied on depth sensors to train their models in simulation and transfer to the real robot.

\vspace{1mm}
\noindent \textbf{Task-Oriented Grasping:} A major portion of research in grasping aims at holding the object in the robot gripper so as to not drop it despite external wrenches.
In practice, however, the end goal of grasping is often to manipulate an object to fulfill a goal-directed task once it has been grasped. When the grasping problem is contextualized in manipulation tasks, a grasp planner that solely satisfies the stability constraints is no longer sufficient to satisfy the task-specific requirements. In classical grasping literature, researchers have developed task-oriented grasp quality measures using task wrench space~\cite{haschke2005task,li1988task,prats2007task}. 

Data-driven approaches have also been used to learn task-related constraints for grasp planning~\cite{dang2012semantic,song2010learning}. These studies incorporate semantic constraints, which specify which object regions to hold or avoid, based on a small dataset of grasp examples. However, these grasping methods cannot entail the success of the downstream manipulation tasks, and the hand-labeled semantic constraints cannot generalize across a large variety of objects. On the contrary, our work jointly learns the task-aware grasping model and the manipulation policy given a grasp. Thus, our grasping model is directly optimized to fulfill its downstream manipulation tasks. Furthermore, we employ deep neural networks to train our task-aware grasping models on a large repository of 3D objects, enabling it to generalize from this repository of objects to unseen objects as well as from simulation to the real world.

\vspace{1mm}
\noindent \textbf{Affordance Learning:} Another line of related work centers around understanding the affordances of objects~\cite{do2017affordancenet,koppula2013learning,zhu2015understanding,zhu2014reasoning}. The notion of affordances introduced by Gibson~\cite{gibson1979affordance} characterizes the functional properties of objects and has been widely used in the robotics community as a framework of reasoning about objects~\cite{katz2014perceiving,varadarajan2012afrob}. 
Prior art has developed methods to learn different forms of object affordance such as semantic labels~\cite{zhu2014reasoning}, spatial maps~\cite{jiang2012learning}, and motion trajectories~\cite{zhu2015understanding}. 
Our work follows a progression of previous work on behavior-grounded affordance learning~\cite{fitzpatrick2003learning,jain2011learning,mar2015self,mar2017self,stoytchev2005behavior}, where the robot learns object affordance by observing the effects of actions performed on the objects. 
Nonetheless, we do not explicitly supervise our model to learn and represent goal-directed object affordances. Instead, we demonstrate that our model's understanding of object affordance naturally emerges from training grasping and manipulation simultaneously. Recent work by \cite{mar2017self} has the closest resemblance to our problem setup; however, their action space consists of a small set of discrete actions, while we employ a multi-dimensional continuous action space. Aside from the problem, their method of self-organizing maps uses hand-designed tool pose and affordance descriptors, while we eschew feature engineering in favor of end-to-end deep learning.


%% file: 3-problem-statement.tex
\section{Problem Statement}
\label{sec:ps}

Our goal is to control a robot arm to perform tool-based manipulation tasks using novel objects. Each task is a two-stage process. In the first stage, the robot grasps an object as a tool for a task. In the second stage, the robot manipulates the tool to interact with the environment to complete the goal of the task. The visual appearance of the tool is provided for the robot to accomplish this.

\noindent \textbf{Notations of Grasping:}
The robot operates in a workspace based on camera observations, where $\mathcal{O}$ denotes the observation space. We consider the grasping problem in the 3D space, where $\mathcal{G}$ denotes the space of possible grasps. Given a pair of observation $\mathbf{o}\in\mathcal{O}$ and grasp $\mathbf{g}\in\mathcal{G}$, let $S_{G}(\mathbf{o}, \mathbf{g})\in\{0,1\}$ denote a binary-valued grasp success metric, where $S_{G}=1$ indicates that the grasp is successful according to the predefined metric. In practice, the underlying sensing and motor noise introduce uncertainty to the execution of a grasp. We measure the robustness of a grasp $Q_{G}(\mathbf{o}, \mathbf{g})$ by the probability of grasp success under uncertainty, 
where $Q_{G}(\mathbf{o}, \mathbf{g})=\Pr(S_{G}=1|\mathbf{o}, \mathbf{g})$. This grasp metric $S_{G}$ is \emph{task-agnostic}, which evaluates the quality of a grasp without grounding to a specific task. As we noted in \secref{rw}, data-driven grasping methods~\cite{mahler2017dex,viereck2017learning} have focused on optimizing \emph{task-agnostic} grasps.

\noindent \textbf{Problem Setup:}
By contrast, we contextualize the grasping problem in tool manipulation tasks. In our setup, the grasping stage is followed by a manipulation stage, where a policy $\pi$ produces actions to interact with the environment once the object is grasped. Intuitively, both the choice of grasps and the manipulation policy play an integral role in the success rate of a task. 

Let $S_{T}(\mathbf{o}, \mathbf{g})\in\{0,1\}$ denote a binary-valued task-specific success metric, where $S_{T}=1$ indicates that the task $T$ is successfully done based on the goal specification. 
Clearly the grasp success is the premise of the task success, i.e., $S_T=1$ entails $S_G=1$. Given a manipulation policy $\pi$ for the task, we measure the robustness $Q^{\pi}_{T}$ of a \emph{task-oriented} grasp by the probability of task success under policy $\pi$, where $Q^\pi_{T}(\mathbf{o}, \mathbf{g})=\Pr(S_{T}=1|\mathbf{o}, \mathbf{g})$. 
Thereafter, the overall learning objective is to train both policies simultaneously such that:
\begin{equation}
\mathbf{g}^{*}, \pi^{*}=\arg\max_{\mathbf{g}, \pi}\,\, Q^\pi_{T}(\mathbf{o},\mathbf{g}).    
\end{equation}
We aim at selecting the optimal grasp $\mathbf{g}^{*}$ that is most likely to lead to to the completion of the task, and at the same time finding the best policy $\pi^{*}$ to perform the task conditioned on a grasp.
In practice, we implement both the grasping policy and the manipulation policy using deep neural networks.
We detail the design of the neural network models and their training procedures in Sec.~\ref{sec:method}.

\begin{figure*}[!t]
\vspace{-10pt}
\centering
  \includegraphics[width=0.95\linewidth]{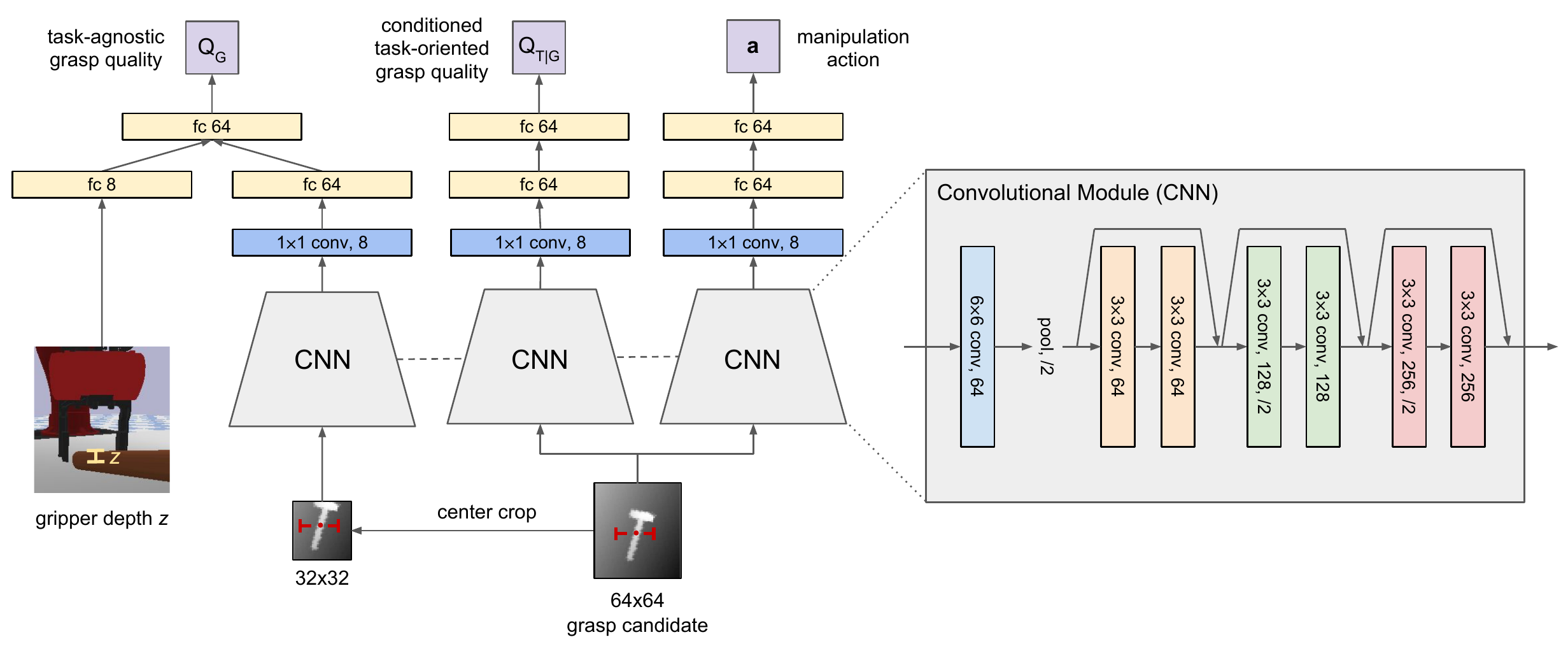}
  \caption{\textbf{Task-Oriented Grasping Network (TOG-Net).} The inputs to the network are two depth image crops and the sampled gripper depth $z$. The network predicts task-agnostic grasp quality, conditioned task-oriented grasp quality, and manipulation actions. The CNN modules share parameters as denoted by the dashed lines. Residual network layers and batch normalization are used in the CNN modules.}
  \label{fig:neural_network}
  \vspace{-15pt}
\end{figure*}

\noindent \textbf{Assumptions:}
We consider the problem of task-oriented grasp planning with a parallel-jaw gripper based on point clouds from a depth camera. The training of our model uses simulated data generated from a real-time physics simulator~\cite{coumans2017bullet}. Our design decision is inspired by the effective use of depth cameras in transferring the grasping model and manipulation policy trained in simulation to reality~\cite{mahler2017dex,viereck2017learning}. Further, to reduce the search space of grasping candidates, we restrict the pose of the gripper to be perpendicular to the table plane. In this case, each grasp $\mathbf{g} = (g_x, g_y, g_z,  g_{\phi})$ has 4 degrees of freedom, where $(g_x, g_y, g_z) \in \mathbb{R}^3$ denotes the position of the gripper center, and $g_{\phi}\in[0,2\pi)$ denotes the rotation of the gripper in the table plane. Each observation $\mathbf{o}\in\mathbb{R}^{H\times W}_{+}$ is represented as a depth image from a fixed overhead RGB-D camera with known intrinsics.

%% file: 4-model.tex
\section{Task-Oriented Grasping for Tool Manipulation} 
\label{sec:method}

As shown in Fig~\ref{fig:overview}, our task-oriented grasping network (TOG-Net) consists of a task-oriented grasping model and a manipulation policy. The two modules are coupled and learn to achieve task success together. In this section, we first present the design and implementation of the two modules, and then describe how they are jointly trained using simulated self-supervision.


\subsection{Learning Task-Oriented Grasp Prediction}

High-quality task-oriented grasps should simultaneously satisfy two types of constraints. First, the tool must be stably held in the robot gripper, which is the goal of task-agnostic grasping. Second, the grasp must satisfy a set of physical and semantic constraints that are specific to each task. 

For a given observation $\mathbf{o} \in \mathcal{O}$, a small subset of grasps $\mathcal{G}_{\alpha} \subseteq \mathcal{G}$ can robustly lift up the object with a grasp quality higher than $\alpha$, i.e. $Q_{G}(\mathbf{o},\mathbf{g}) \geq \alpha$. Hence, the task agnostic grasp prediction problem involves finding the corresponding grasp $\mathbf{g}$ that maximizes the grasp quality $Q_{G}(\mathbf{o},\mathbf{g}) = {\Pr}(S_{G}=1 | \mathbf{o},\mathbf{g})$.
This prediction problem can be solved with a variety of methods, and we build upon the approach of Mahler et al.~\cite{mahler2017dex} that uses quality function approximation with algorithmic supervision via analytic models. 

As noted in \secref{ps}, the overall objective is maximizing the probability of task success $Q_T^{\pi}(\mathbf{o},\mathbf{g})$ under a policy $\pi$, grasp $\mathbf{g}\in \mathcal{G}$. However, directly solving this problem results in a discrete space search over a large space of grasps, and then a subsequent optimization problem to solve for a manipulation policy given each grasp. Furthermore, we note that tool manipulation task execution only succeeds if the grasp has succeeded. However, not all grasps $\mathbf{g} \in \mathcal{G}_{\alpha}$ result in a successful task. Specifically, we can define a conditional robustness metric $Q_{T|G}$ that measures the probability of task success (under policy $\pi$) conditioned on a successful grasp, where $Q_{T|G}(\mathbf{o},\mathbf{g}) = \Pr_\pi(S_{T}=1|S_{G}=1,\mathbf{o},\mathbf{g})$. Then, task-oriented grasps form a subset of task-agnostic grasps: $\mathcal{G}_{\alpha,\delta} \subseteq \mathcal{G}_{\alpha}$, i.e. the grasps which lead to task success with a task quality conditioned on the grasp $Q_{T|G}(\mathbf{o},\mathbf{g}) \geq \delta$ where  $\delta$ is a chosen threshold.

This formulation lets us to effectively decouple the two problems as: 1) finding robust task-agnostic grasps; and 2) finding a robust task-oriented grasp among robust grasps and the corresponding manipulation policy. The key observation is that the task quality metric $Q_T^{\pi}(\mathbf{o},\mathbf{g})$ can be factorized into independently computable terms: $Q_{T|G}(\mathbf{o},\mathbf{g})$ and $Q_{G}(\mathbf{o},\mathbf{g})$ . Formally, the task robustness $Q_{T}(\mathbf{o},\mathbf{g})$ can be decomposed as follow:
\begin{align*}
Q_{T}^{\pi}(\mathbf{o},\mathbf{g}) & = {\Pr}_\pi(S_{T}=1|\mathbf{o},\mathbf{g}) \\
& = {\Pr}_\pi(S_{T}=1,S_{G}=1|\mathbf{o},\mathbf{g})\\
& = {\Pr}_\pi(S_{T}=1|S_{G}=1,\mathbf{o},\mathbf{g}) \cdot {\Pr}(S_{G}=1|\mathbf{o},\mathbf{g}) \\
& = Q_{T|G}(\mathbf{o},\mathbf{g}) \cdot Q_{G}(\mathbf{o},\mathbf{g}).
\end{align*}

Our model learns to approximate the values of grasp quality $Q_{G}(\mathbf{o},\mathbf{g})$ and task quality conditioned on grasp $Q_{T|G}(\mathbf{o},\mathbf{g})$ using deep neural networks given object $\mathbf{o}$ and grasp $\mathbf{g}$ as inputs. We denote the predicted values as $\hat{Q}_{G}(\mathbf{o},\mathbf{g}; \theta_1)$ and $\hat{Q}_{T|G}(\mathbf{o},\mathbf{g}; \theta_2)$, where $\theta_1$ and $\theta_2$ represent the neural network parameters.

The pipeline during testing is shown in \figref{overview}. We first sample 200 antipodal grasp candidates based on depth gradients~\cite{mahler2017dex}. Then $\hat{Q}_{T}(\mathbf{o},\mathbf{g}; \theta_1, \theta_2) = \hat{Q}_{T|G}(\mathbf{o},\mathbf{g}; \theta_2) \cdot \hat{Q}_{G}(\mathbf{o}, \mathbf{g}; \theta_1)$ is computed for each grasp candidate. We run the cross-entropy method~\cite{rubinstein2004cross} for 3 iterations as in \cite{mahler2017dex}, in order to rank and choose the task-oriented grasp corresponds to the highest $\hat{Q}_{T}$.

\subsection{Learning the Manipulation Policy}
\label{sec:manipulation_policy}


To complete the task with different tools and different grasps, the manipulation policy needs to be conditioned on $\mathbf{o}$ and $\mathbf{g}$. The manipulation policy can be either an external motion planner or a learned policy. While our pipeline is not limited to a specific action space, here we choose to use parameterized motion primitives parallel to the planar table surface to control the robot in an open-loop manner. After the motion primitive is chosen based on the task environment, our manipulation policy predicts the continuous manipulation actions $\mathbf{a} = (a_x, a_y, a_z, a_\phi) \in \mathbb{R}^3$, where $(a_x, a_y, a_z)$ and $a_\phi$ are the translation and rotation of the motion primitive. We use a Gaussian policy $\pi(\mathbf{a} | \mathbf{o}, \mathbf{g}; \theta_3) = \mathcal{N}(f(\mathbf{o}, \mathbf{g}; \theta_3), \Sigma)$, where $f(\mathbf{o}, \mathbf{g}; \theta_3)$ is a neural network for predicting the mean with parameters $\theta_3$ and the covariance matrix $\Sigma$ is a constant diagonal matrix.

\subsection{Neural Network Architecture}

In \figref{neural_network} we propose a three-stream neural network architecture for jointly predicting $\hat{Q}_{G}$ and $\hat{Q}_{T|G}$ and $\mathbf{a}$. Following the practice of \cite{mahler2017dex}, we convert $\mathbf{o}$ and $\mathbf{g}$ into gripper depth $z$ and image crops as inputs to the neural network. The gripper depth is defined as the distance from the center of the two fingertips to the object surface. The image crops are centered at the grasp center $(g_x, g_y, g_z)$ and aligned with the grasp axis orientation $\phi$. \cite{mahler2017dex} uses image crops of size $32 \times 32$ to focus on the contact between the gripper and the tool. To achieve the task success, our model is supposed to reason about the interactions between the tool and the task environment which requires a holistic understanding of the shape of the tool. Thus our model predicts $\hat{Q}_{T|G}$ and $\mathbf{a}$ using larger image crops of size $64 \times 64$ which covers most of training and testing objects. Meanwhile the center crop of $32 \times 32$ is used to predict $\hat{Q}_{G}$. Our neural network is composed of three streams which share parameters in their low-level convolutional layers, extracting identical image features, denoted by dotted lines. Building atop the GQCNN in \cite{mahler2017dex}, we use residual network layers~\cite{he2016deep} and batch normalization~\cite{ioffe2015batch} to facilitate the learning process. On top of the convolutional layers with shared weights, we apply bottleneck layers of $1 \times 1$ convolutional filters for each stream to reduce the size of the network.

\subsection{Learning Objectives and Optimization}

We jointly train the task-oriented grasping model and the manipulation policy with simulated robot experiments of grasping and manipulation. Each simulated episode in our training dataset contains sampled grasp $\mathbf{g}$, action $\mathbf{a}$ and the resultant grasp success label $S_{G}$, task success label $S_{T}$. We use cross-entropy loss $\mathcal{L}$ for training the grasp prediction functions $\hat{Q}_{G}$ and $\hat{Q}_{T|G}$. For training the policy $\pi$, we use the policy gradient algorithm with gradients $\nabla \log \pi(\mathbf{a} | \mathbf{o}, \mathbf{g}; \mathbf{\theta_3})$. We use the task success label as the reward of the manipulation policy. Since we are using a Gaussian policy as described in Sec.~\ref{sec:manipulation_policy}, this is equivalent to minimizing $\frac{1}{2}||f(\mathbf{o}, \mathbf{g}; \theta_3) - \mathbf{a}||^2_{\Sigma} \cdot S_{G}$ with respect to $\theta_3$. Let the parameters of the neural network to be denoted as $\theta = \{\theta_1, \theta_2, \theta_3\}$, we jointly train our model by solving the following optimization problem:
\begin{equation}
\begin{split}
    \mathbf{\theta}^* = \arg\min_{\mathbf{\theta}} \sum_{i=1}^{N} & \mathcal{L}(S_{G}, \hat{Q}_{G}(\mathbf{o}, \mathbf{g}; \theta_1)) \\
    &+ \mathds{1}[S_{G}=1] \cdot \mathcal{L}(S_{T},\hat{Q}_{T|G}(\mathbf{o}, \mathbf{g}; \theta_2)) \\
    &+ \mathds{1}[S_{T}=1] \cdot \frac{1}{2}||f(\mathbf{o}, \mathbf{g}; \theta_3) - \mathbf{a}||^2_{\Sigma}.
\end{split}
\end{equation}
where $\mathds{1}(\cdot)$ is the indicator function.

%% file: 5-dataset.tex
\section{Self-Supervision for Grasping and Manipulation}

\begin{figure*}[!ht]
\vspace{-5pt}
\centering
  \includegraphics[width=1.0\linewidth]{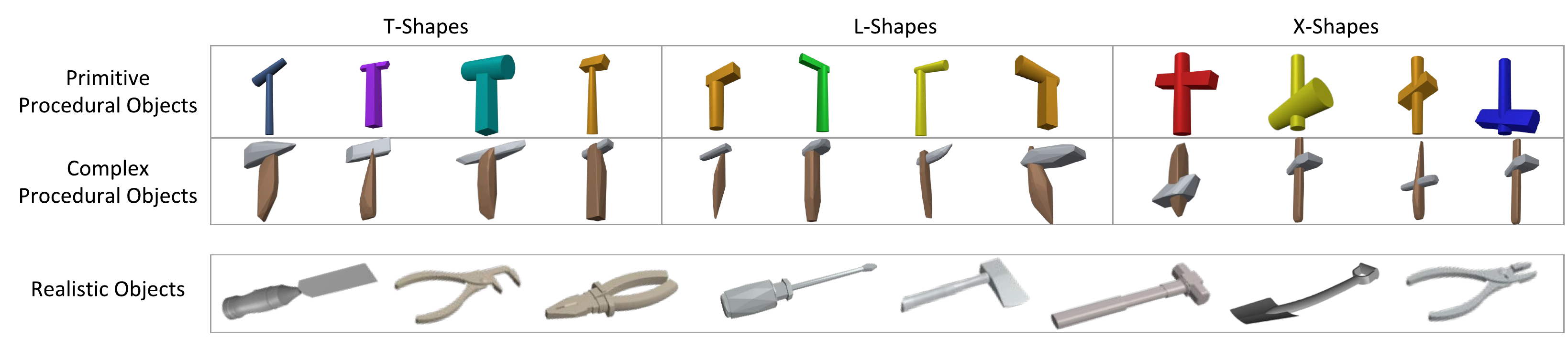}
  \caption{\textbf{Example of objects in simulation.} The first two rows show the procedurally generated objects based on shape primitives as well as complex meshes. These objects are generated using three predefined composing rules to result in T-shapes, L-shapes, and X-shapes. The last row shows the realistic shapes from existing 3D model datasets. }
  \label{fig:procObj}
  \vspace{-6mm}
\end{figure*}

\subsection{Procedural Generation of Tool Objects}
We train our model in simulation with a large repository of 3D models so as to generalize to unseen objects. However, existing 3D model datasets do not contain enough objects suitable for tool manipulation while exhibiting rich variations in terms of their geometric and physical properties. As shown in \figref{procObj}, we leverage a common strategy of procedural generation~\cite{bousmalis2017using,tobin2017randomization} to produce a large set of diverse and realistic objects that can be used as tools for tasks we are interested in.

While the generation process can be arbitrarily complex, we choose to generate objects composed of two convex parts. The two parts are connected by a fixed joint. We define three type of composed shapes: T-shapes, L-shapes, and X-shapes. For each object, two convex meshes are first sampled. Then the meshes are randomly scaled along the x, y, and z axes. Depending on the type of object shape, the parts are shifted and rotated with respect to each other. We randomly sample physical dynamic properties such as density and friction coefficients.

We use two sets of meshes to generate two different set of objects: primitive and complex. We generate primitive meshes that are composed by a set of parameterized shape primitives including cuboids, cylinders, and polytopes. The dimensions and textures are randomly chosen from predefined ranges. The primitive meshes are generated by OpenScad \cite{openscad}. We also obtain convex object parts from a variety of realistic 3D object models as the complex meshes. This done by running convex decomposition~\cite{mamou2009simple} on each object from \cite{bohg2014data}.
\vspace{-1mm}

\subsection{Data Generation with Simulated Self-Supervision}

In order to collect large-scale datasets for training and evaluating our model, we develop a self-supervision framework to automatically generate training data. We leverage an open-source real-time physics simulator, Bullet~\cite{coumans2017bullet}, which allows a simulated robot to perform trial and error in millions of trails. We record grasp and task success labels in each trial and use them to train our models described in \secref{method}. For training each task we collect the data in three rounds. After each round, we train the grasping model and the manipulation policy using the collected data to obtain an updated model. In each round we run the simulation for 500K trials.

In the first round, we perform a random policy using a GQCNN model trained on Dex-Net 2.0 Dataset~\cite{mahler2017dex}. The original GQCNN model uses a cross entropy method~(CEM)~\cite{rubinstein2004cross} to sample robust grasps corresponds to the highest task-agnostic grasp quality scores. But the trained GQCNN usually lead to a collapsed mode of grasps which is most robust according to the ranking of the predicted scores. Ideally we want to collect data of diverse sampled grasps and evaluate how well they can be used in each task. An alternative is to sample uniformly sample grasps with grasp quality scores higher than a threshold. In practice we found such sampling usually clusters on the long edge of a tool object since there are more antipodal grasps possible there. To encourage diverse exploration, we instead use non-maximum suppression~(NMS)~\cite{hartley2003multiple} which is widely used in object detection algorithms. The NMS algorithm goes through the ranked antipodal grasps, and removes grasps which have short Euclidean distances with previous grasps with higher grasp quality scores. This guarantees all remaining grasps are separate from each other and usually produces 10 to 30 distinguished modes. With these sampled grasp, the random policy uniformly samples manipulation action from the action space for each task. 

In the following rounds, we use the $\epsilon$-greedy strategy with the updated grasping model. The grasping model uses the CEM method described in Section~\ref{sec:method} with probability $1 - \epsilon_1$, and uses the NMS method with GQCNN predictions as described above with $\epsilon_1$ probability. The manipulation policy predicts and manipulation action parameters in Section~\ref{sec:method} with probability $1 - \epsilon_2$, and use random actions with probability $\epsilon_2$. We set 0.2 for both $\epsilon_1$ and $\epsilon_2$. 

%% file: 6-experiment.tex
\section{Experiments}
\label{sec:expt}

The goal of our experimental evaluation to answer following questions: 
(1) Does our method improve task performance as compared to baseline methods? 
(2) Does the joint training qualitatively change the mode of grasping? 
(3) Can the model be trained with simulated self-supervision work in the real world?

We evaluate our method on two tabletop manipulation tasks: \textit{sweeping} and \textit{hammering}. We define our hammering task as a motion primitive that achieves fitting a peg in a hole with tight tolerance, which is prevalent in assembly tasks as shown in \cite{williamson1999robot, komizunai2008experiments}. Sweeping, on the other hand, is a primitive in autonomous manipulation, such as in part positioning and reorientation~\cite{lynch1996stable}, grasping in clutter~\cite{dogar2011framework}, object manipulation without lifting~\cite{mericcli2015push}. Sweeping with tools has been studied in the context of singulation and part retrieval~\cite{eitel2017learning,laskey2017comparing}. Each of these tasks requires grasping objects in specific modes which can often be different from the best stable grasp available, thereby resulting in competing objectives.  

We evaluate our model in both simulation and the real world. The basic setup of both tasks includes a 7-DoF Rethink Robotics Sawyer Arm with a parallel jaw gripper, a $48'' \times 30''$ table surface, and an overhead Kinect2 camera. In simulation, the robot and camera are placed according to the real-world camera calibration results, in order to obtain consistent performance. For both experiments, we use models solely trained using simulated data as described in \secref{method}.




\begin{figure*}[t]
    \vspace{-5pt}
    \centering
    \begin{subfigure}{0.5\textwidth}
        \centering
        \includegraphics[width=\textwidth]{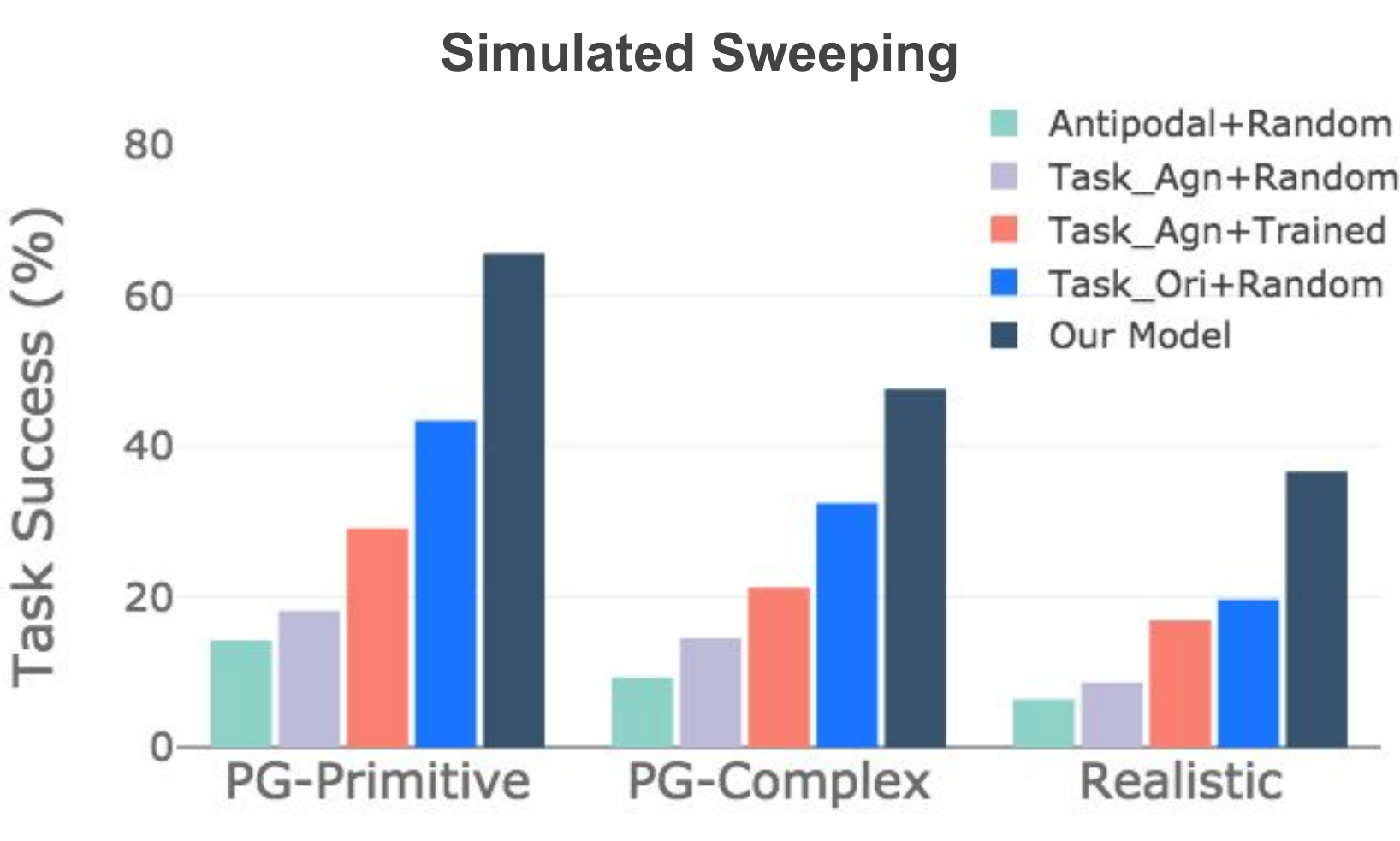}
    \end{subfigure}%
    \begin{subfigure}{0.5\textwidth}
        \centering
        \includegraphics[width=\textwidth]{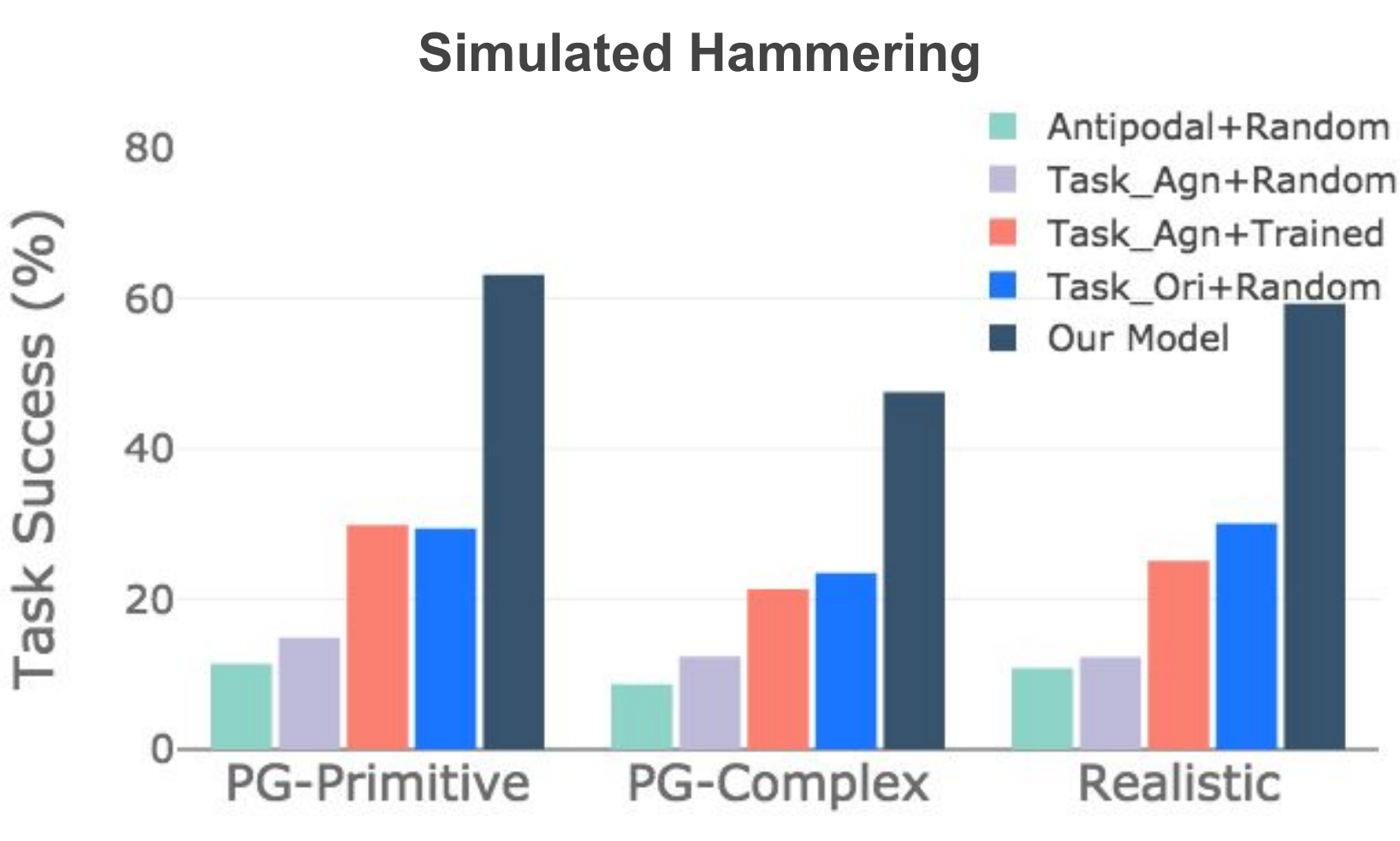}
    \end{subfigure}%
    \caption{\textbf{Performance of simulated experiments.} We perform an evaluation of our model for sweeping and hammering in simulation. We compare performance separately on three object categories as shown in \figref{procObj}: procedurally generated objects with primitive meshes (PG-Primitive), procedurally generated objects with complex meshes (PG-Complex), and 3D objects from existing datasets (Realistic). Our model outperforms all baselines using the three object categories in both tasks.}
    \label{fig:sim-results}
    \vspace{-10pt}
\end{figure*}

\subsection{Task Design}
The table surface is split into two halves: a grasping region and a manipulation region. Before each episode starts, an object is sampled and randomly dropped onto the grasping region to be used as the tool. A depth image is taken from the overhead Kinect2 camera as the input of the model. The model then predicts the 4-DOF grasp and the parameters of the motion primitive. The robot grasps the object from the grasping region and performs the task in the manipulation region. In our task design, the motion primitive is a predefined single step action. Our model predicts the starting gripper pose relative to a reference point.

\noindent \textbf{Sweeping:} Target objects are randomly placed in the manipulation region as the target objects. In the real world we use two soda cans as the target objects, and in simulation we randomly place one or two 3D models of cans. The task goal is to sweep all target objects off the table using the tool. The motion primitive of sweeping is a straight line trajectory of the gripper parallel to the table surface. The gripper trajectory starts from the pose $(a_x, a_y, a_z, a_{\phi})$ and moves 40cm along y-axis of the world frame. $(a_x, a_y, a_z, a_{\phi})$ is predicted relative to the mean position of the target objects. The task success is achieved when all target objects contact the ground. For robust sweeping, the tool ideally need to have a large flat surface in contact with the target object.

\noindent \textbf{Hammering:} A peg and a slot are randomly placed in the manipulation region, where the peg is horizontally half-way inserted into the slot. The task goal is to use the tool to hammer the peg fully into the slot. The motion primitive of hammering is a rotation of the gripper along the z-axis. The trajectory starts with the gripper pose $(a_x, a_y, a_z, a_{\phi})$ and ends after the last arm joint rotates by 90 degree counterclockwise at full speed. $(a_x, a_y, a_z, a_{\phi})$ is predicted relative to the position of the peg. The task success is achieved when the whole peg is inside the slot. This task requires a sufficient contact force between the tool and the peg to overcome the resistance. Meanwhile the tool should avoid collisions with the peg before the hammering.

\vspace{-2mm}
\subsection{Experiment Setup}

Training uses 18,000 procedurally generated objects including 9,000 PG-Primitive objects and 9,000 PG-Complex objects. In addition to randomizing physical properties, we randomly sample the camera pose and intrinsics by adding disturbances to the values obtained from the real-world setup. 

During testing, we use 3000 instances of each type of procedurally generated object. We also test on 55 realistic objects selected from Dex-Net 1.0~\cite{mahler2016dex} and MPI Grasping dataset~\cite{bohg2014data}. These objects contain both tool-like and non-tool like objects as shown in \figref{procObj}. None of these test objects are seen during training. 


We compare our method to 4 baselines:
\begin{enumerate}
  \item \textit{Antipodal+Random}: Use a sampled antipodal grasp with a random action uniformly sampled with $x, y, z$ positions in $[-5, 5]$  in terms of centimeters and $\theta$ in $[-\frac{\pi}{20}, \frac{\pi}{20}]$.
  \item \textit{Task\_Agn+Random}: A task-agnostic grasp from Dex-Net 2.0~\cite{mahler2017dex} with a random action.
  \item \textit{Task\_Agn+Trained}: Same as above but with a manipulation policy trained with our method. This is akin to the current best solution.
  \item \textit{Task\_Ori+Random}: An ablative version of our model with task-oriented grasps executed with a randomized action. 
\end{enumerate}

\subsection{Simulated Experiments}
We evaluate our method on both tasks using the simulated setup described above. For each algorithmic method, we run 100,000 episodes in simulation and report the task success rate. The task success analysis for each of the baselines and our method is presented in \figref{sim-results}. 

Our model outperforms the four baselines in both tasks for all object categories. The contrast is more significant for hammering than sweeping. This is because hammering requires well-trained manipulation policy to direct the tool to hit the peg. A small deviance from the optimal hammering trajectory can let the tool miss the peg or collide with the slot. While for the sweeping task, when the robot uses a long edge of the tool to sweep, there is a high tolerance of manipulation action errors. Even random actions can often succeed. Among the three object categories, PG-Primitive is usually the easiest to manipulate with. Complex meshes cause more grasping failures and are harder for their geometric properties are harder to reason about. Realistic objects are not usually good for sweeping since they are more roundish and very few have long edges. While the hammering performance with realistic objects are much better, because many of these objects are cylinder objects with a bulky head and even actual hammers. 


\subsection{Real-World Experiments}

\begin{figure}[!t]
\centering
  \includegraphics[width=0.85\linewidth]{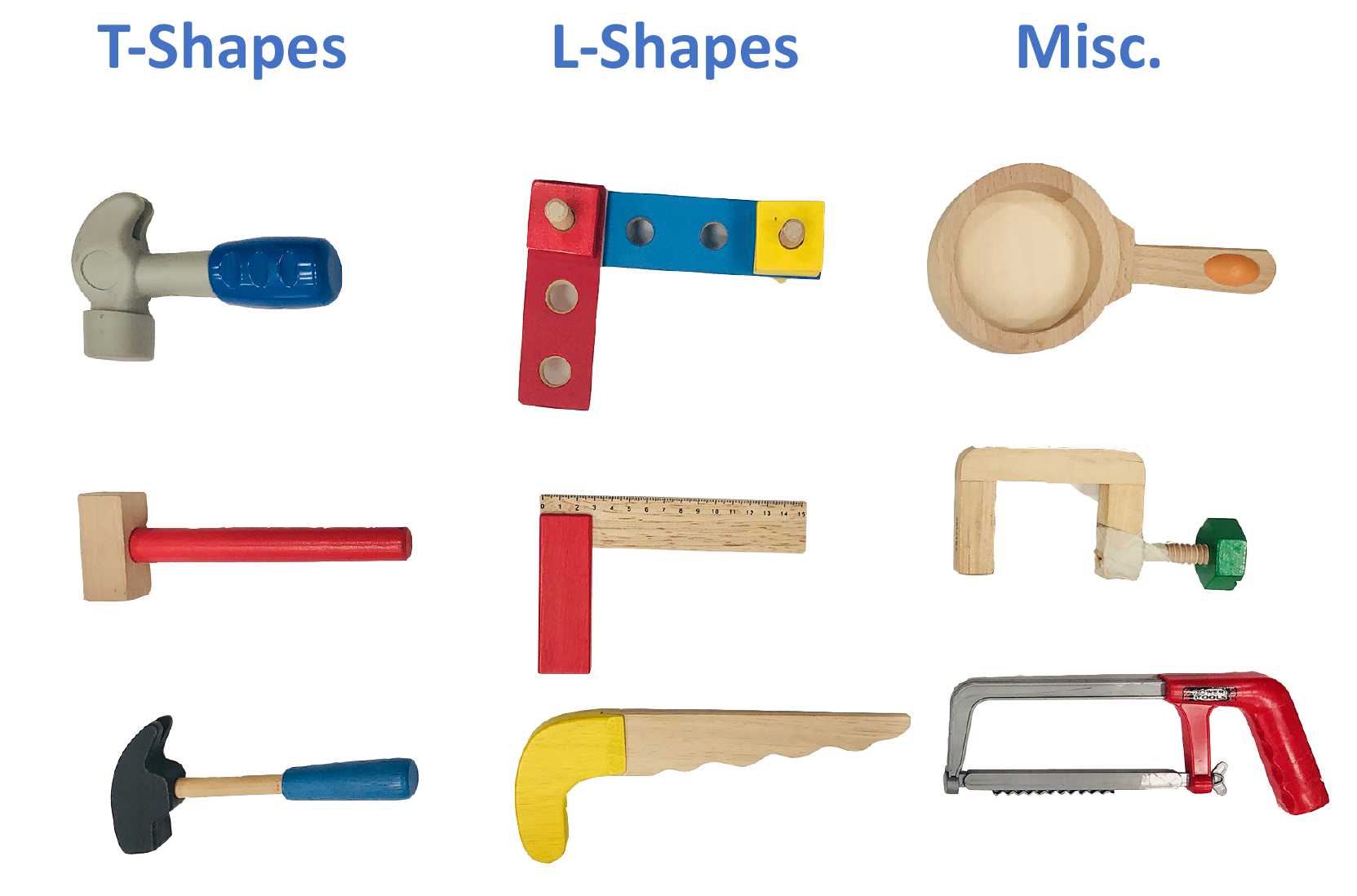}
  \caption{\textbf{Real-world test objects.} We used 9 unseen objects for our real-world experiments. These objects are grouped into three categories: T-shapes, L-shapes and miscellaneous objects.}
  \label{fig:objects-real}
\end{figure}

For our real-world experiments, we use 9 unseen objects consisting of three categories as shown in \figref{objects-real}. T-shape and L-shape objects have similar geometric properties with our procedurally generated objects during training, whereas the miscellaneous objects have structures and curvatures totally unseen during training. 

In the real world, we compare our model with two baseline methods: antipodal grasping with trained manipulation policy (antipodal + trained) and task-agnostic grasping with trained manipulation policy (task-agnostic + trained). We perform each task with each object for 5 robot trials for a total of 270 trials. The per-category and overall task success rates are shown in \tabref{robot-expt}. For all object categories, our model achieved better performance compared to the baselines.

For sweeping, our model can successfully grasp the head of T-shapes or the short edge of L-shapes, and sweep with the longer part. For more complex miscellaneous objects, it is less obvious for the model to figure out which part should be grasped. But for most trials, the grasp predicted by our model is intuitive to humans and leads to successful sweeping. For T-shapes, the difference between task-agnostic and task-oriented grasping is larger since the object usually only has one long handle. In contrast for some L-shapes, the two edges are both long enough for the task, grasping either edge does not make a significant difference. For miscellaneous objects, the model can have problems reasoning about novel object parts. For instance, it sometimes chooses to grasp the handle of the pan and sweep with the round part, which is unstable for sweeping roundish target objects.

For hammering, our model performs equally well for T-shapes and L-shapes. And the failures are usually caused by occasional deviations during the execution of grasping or manipulation. As a comparison, baseline models often choose to grasp the head and hammer with the handle, which is sub-optimal. Compared to T-shapes and L-shapes, there might not be an obvious way to use miscellaneous objects as hammers. Among miscellaneous objects, the pan can be used as a hammer very well. The model tends to grasp the handle and hammer with the bulky roundish part. 


\vspace{-4mm}

\begin{table}[!t]
\begin{minipage}[t]{\linewidth}
    \centering
    \begin{tabular}{l|c|c|c}
\hline
 & \multicolumn{3}{c}{\cellcolor[HTML]{FFCC67}\textbf{Grasping Model}} \\ \cline{2-4} 
\multirow{-2}{*}{\textbf{\begin{tabular}[c]{@{}l@{}}Real-World \\ Sweeping\end{tabular}}} & \multicolumn{1}{c|}{\cellcolor[HTML]{CBCEFB}\textbf{\begin{tabular}[c]{@{}c@{}}Antipodal \\ + Trained\end{tabular}}} & \multicolumn{1}{c|}{\cellcolor[HTML]{CBCEFB}\textbf{\begin{tabular}[c]{@{}c@{}}Task-Agnostic\\ + Trained\end{tabular}}} & \multicolumn{1}{c}{\cellcolor[HTML]{CBCEFB}\textbf{Our Model}} \\ \hline
\cellcolor[HTML]{FFCC67}T-Shapes & 13.3  & 20.0 & \textbf{73.3} \\
\cellcolor[HTML]{FFCC67}L-Shapes & 23.7 & 46.7 & \textbf{80.0} \\
\cellcolor[HTML]{FFCC67}Misc & 33.3 & 13.3 & \textbf{60.0} \\ \hline
\rowcolor[HTML]{E0E0E0} 
\textbf{Overall} & 24.4 & 23.6 & \textbf{71.1}   \\ \hline
\end{tabular}
\end{minipage}

\vspace{4mm}

\centering

\begin{minipage}[t]{\linewidth}
    \centering
    \begin{tabular}{l|c|c|c}
\hline
 & \multicolumn{3}{c}{\cellcolor[HTML]{FFCC67}\textbf{Grasping Model}} \\ \cline{2-4} 
\multirow{-2}{*}{\textbf{\begin{tabular}[c]{@{}l@{}}Real-World \\ Hammering\end{tabular}}} & \multicolumn{1}{c|}{\cellcolor[HTML]{CBCEFB}\textbf{\begin{tabular}[c]{@{}c@{}}Antipodal \\ + Trained\end{tabular}}} & \multicolumn{1}{c|}{\cellcolor[HTML]{CBCEFB}\textbf{\begin{tabular}[c]{@{}c@{}}Task-Agnostic\\ + Trained\end{tabular}}} & \multicolumn{1}{c}{\cellcolor[HTML]{CBCEFB}\textbf{Our Model}} \\ \hline
\cellcolor[HTML]{FFCC67}T-Shapes &  46.7 & 60.0 & \textbf{86.7} \\
\cellcolor[HTML]{FFCC67}L-Shapes & 13.3  & 33.3 & \textbf{86.7} \\
\cellcolor[HTML]{FFCC67}Misc & 40.0 & 53.3 & \textbf{66.7} \\\hline
\rowcolor[HTML]{E0E0E0} 
\textbf{Overall} & 33.3 & 44.4 & \textbf{80.0}  \\ \hline
\end{tabular}
\end{minipage}%

\vspace{2mm}

\caption{\textbf{Performance of real-world experiments.} We compare our model with other grasping methods in terms of task success rates. We use 9 real-world objects grouped into 3 categories. We perform 5 trials with each object for each method, for a total of 270 robot trials. The per-task and overall task success rates are reported in each cell.} 
\label{tab:robot-expt}


\end{table}

\subsection{Qualitative Analysis}
\begin{figure}[!t]
\centering
  \includegraphics[width=\linewidth]{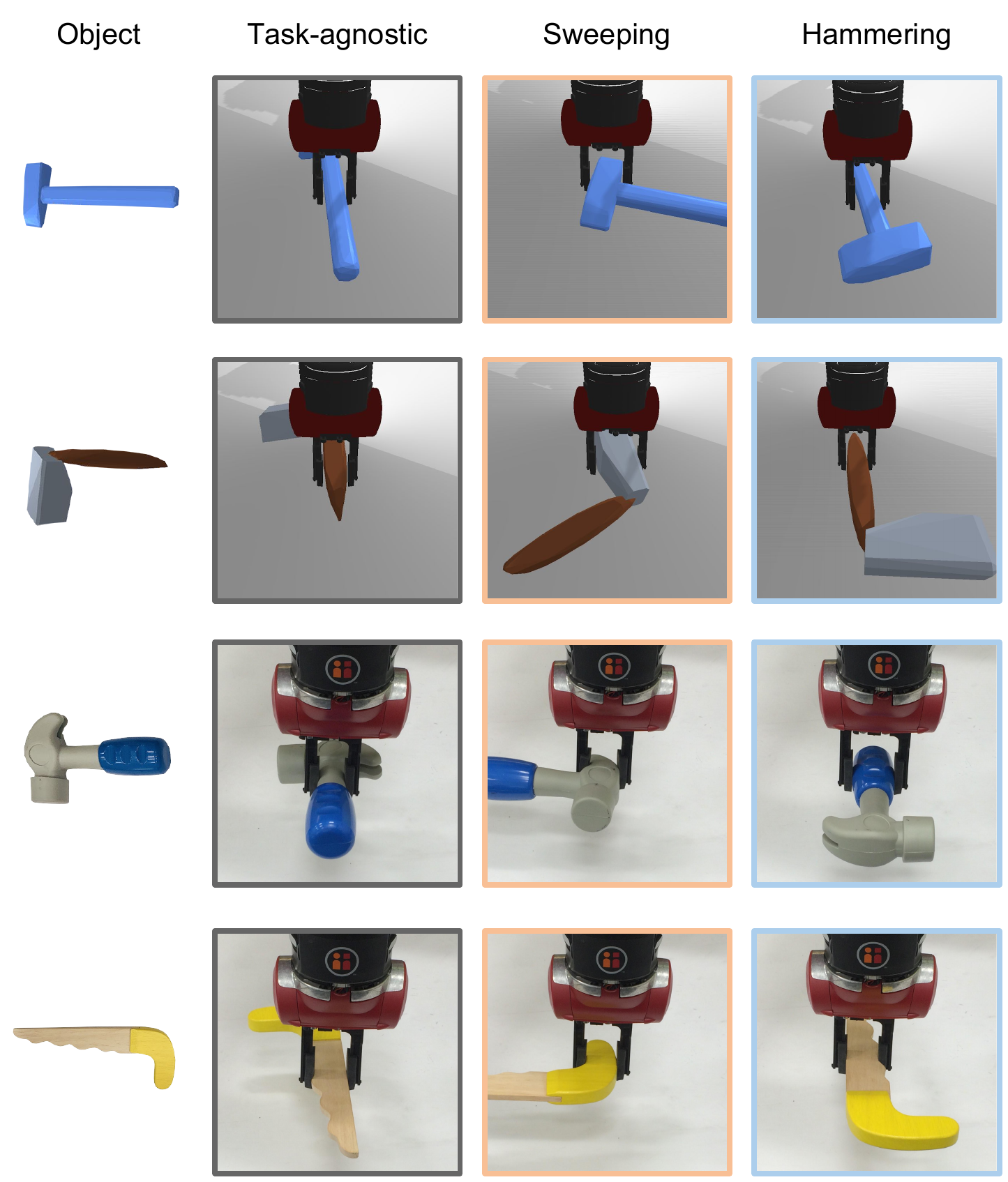}
  \caption{\textbf{Qualitative results.} Column 1 shows RGB images of the tool objects. Column 2 shows example task-agnostic grasps. And Columns 3 and 4 show task-oriented grasps chosen by our model for the sweeping and hammering tasks. Our model favors wide flat surfaces for sweeping and long moment arms for hammering.}
  \label{fig:qualitative}
\end{figure}

In \figref{qualitative}, we demonstrate the same object can be grasped for different tasks by our trained model. Here we show the modes of task-agnostic grasping and task-oriented grasping for 4 example objects, 2 in simulation and 2 in the real world. 

For the sweeping task, it is challenging to sweep all target objects off the table in one shot. It requires the tool to have a flat contact surface to facilitate the manipulation of roundish objects and to sweep across large enough area to catch both of them. Our model learns to grasp the end of the tool object and spare as much surface area as possible for sweeping. This enables the robot to robustly sweep the cans most of the time.

For the hammering task, the main concerns are overcoming the resistance of the peg while avoiding collisions during hammering. We expect the robot to grasp the far end of the handle and hit the peg with the bulky part as the hammer head. Ideally, this could generate the largest torque on the hammer head when hitting the peg. In practice, we found the trained model tends to grasp a little closer to the hammer head on the handle. This is because we use a parallel jaw gripper and it is hard to balance the tool when the fingers are far away from the center of mass. 

For robust task-agnostic grasping, the GQCNN model usually chooses the thin part near the center of mass. Although this sometimes still overlaps with the set of task-oriented grasps, it is not guaranteed if the selected grasp from the task-agnostic GQCNN is suitable for the task. In \figref{qualitative} we show example task-agnostic grasps which are different from the task-oriented grasps mentioned above. 


%% file: 7-discussion.tex
\section{Conclusion} 
\label{sec:conclusion}
We develop a learning-based approach for task-oriented grasping for tool-based manipulation trained using simulated self-supervision. It jointly optimizes a task-oriented grasping model and its accompanying manipulation policy to maximize the task success rate. We leverage a physics simulator that allows a robot to autonomously perform millions of grasping and manipulation trials. The trial and error of the robot provides training data to supervise the deep neural network models. Our experimental results demonstrate that the task-oriented grasps selected by our model are more suitable for downstream manipulation tasks than the task-agnostic grasps.

In the future, our goal is to further improve the effectiveness and robustness of our model by training on a large dataset of realistic 3D models. Additionally, we plan to scale up our model to complex manipulation tasks with end-to-end trained closed-loop manipulation policies. Supplementary material is available at: \href{http://bit.ly/task-oriented-grasp}{bit.ly/task-oriented-grasp}

%% file: 8-acknowledgement.tex
\section*{Acknowledgement}
\label{sec:acknowledgement}
We acknowledge the support of Toyota (1186781-31-UDARO). We thank Ozan Sener for constructive discussions about the problem formulation. We thank Alex Fu, Julian Gao and Danfei Xu for helping with the real-world infrastructure. We thank Erwin Coumans for helping with the Bullet simulator.